\documentclass[letterpaper, 10 pt, conference]{ieeeconf}  % Comment this line out if you need a4paper
\IEEEoverridecommandlockouts                             
\usepackage[T1]{fontenc}
\usepackage{graphicx}
\usepackage{makecell}
\usepackage{multirow}
\usepackage{threeparttable}  
\usepackage{float}
\usepackage{xltabular}
\usepackage{algorithm, algorithmic}
\usepackage{amsmath} % assumes amsmath package installed
\usepackage{amssymb}  % assumes amsmath package installed

\usepackage{subcaption} 
\usepackage[font=small,labelfont=bf,tableposition=top,hypcap=false]{caption}
\usepackage[export]{adjustbox} % for valign=m
\usepackage{helvet}         % selects Helvetica as sans-serif font
\usepackage{courier}        % selects Courier as typewriter font
\usepackage{type1cm}        % activate if the above 3 fonts are
\usepackage{makeidx}         % allows index generation
\usepackage{bm}
\usepackage{notoccite}%make refs turn up in ordeFr

\newcommand{\etal}{\textit{et al.~}}
  
\def\eg{\textit{e.g.},~}

\usepackage{array}

\makeatletter
 \let\NAT@parse\undefined
 \makeatother
\usepackage[pdftex,
colorlinks=true,
bookmarks=true,
citecolor=red,
linkcolor=blue,
pagebackref=false,
pdfstartview=Fit,
breaklinks=true
]{hyperref}

\title{\LARGE \bf
PoseFusion2: Simultaneous Background Reconstruction and Human Shape Recovery in Real-time
}

\author{Huayan Zhang$^{1}$, Tianwei Zhang$^{1,*}$, Tin Lun Lam$^{2,1*}$, and Sethu Vijayakumar$^{3,4}$% <-this % stops a spaced

\thanks{$^{1}$Shenzhen Institute of Artificial Intelligence and Robotics for Society, the Chinese University of Hong Kong, Shenzhen.}%
\thanks{$^{2}$School of Science and Engineering, the Chinese University of Hong Kong, Shenzhen.}
\thanks{$^{3}$The School of Informatics, University of Edinburgh, Edinburgh and The Alan Turing Institute, UK.}
\thanks{$^{4}$The author is a visiting researcher with the Shenzhen Institute of Artificial Intelligence and Robotics for Society (AIRS).}%
\thanks{$^*$Corresponding Author:  \tt\small zhangtianwei@cuhk.edu.cn; \tt\small tllam@cuhk.edu.cn}
}

\begin{document}
\maketitle
\thispagestyle{empty}
\pagestyle{empty}

%%%%%%%%%%%%%%%%%%%%%%%%%%%%%%%%%%%%%%%%%%%%%%%%%%%%%%%%%%%%%%%%%%%%%%%%%%%%%%%%
\begin{abstract}
%圈定workspace：
%动态slam
%多人体环境
%提出困难问题，和我们的解决方案
%最后说一下我们目前达到的效果
Dynamic environments that include unstructured moving objects pose a hard problem for Simultaneous Localization and Mapping (SLAM) performance. 
The motion of rigid objects can be typically tracked by exploiting their texture and geometric features. However, humans moving in the scene are often one of the most important, interactive targets -- they are very hard to track and reconstruct robustly due to non-rigid shapes.
In this work, we present a fast, learning-based human object detector to isolate the dynamic human objects and realise a real-time dense background reconstruction framework. We go further by estimating and reconstructing the human pose and shape. The final output environment maps not only provide the dense static backgrounds but also contain the dynamic human meshes and their trajectories. Our Dynamic SLAM system runs at around 26 frames per second (\emph{fps}) on GPUs, while additionally turning on accurate human pose estimation can be executed at up to 10 \emph{fps}. 
\end{abstract}

%%%%%%%%%%%%%%%%%%%%%%%%%%%%%%%%%%%%%%%%%%%%%%%%%%%%%%%%%%%%%%%%%%%%%%%%%%%%%%%%
\section{Introduction}
%提出困难问题的具体原因，描述，对标工作
%和pf1的区别%
Increasingly, robots are expected to work in close collaboration with humans in dynamic environments. Such collaborative working spaces may contain multiple types of dynamic objects that invalidate classical SLAM frameworks.
% These dynamic objects include: 
% (a) Interactive objects, \eg doors, tools, cups, \etc 
% (b) Rigid moving objects, \eg cars, other mobile robots, \etc and 
% (c) Non-rigid dynamic objects, \eg humans and animals.
Many recent works \cite{pf,DynaSLAM, sf, mvo} have addressed the problem by attempting to identify and remove the dynamic objects, in order to apply the classical static SLAM frameworks. 
These methods directly remove dynamic objects to improve the robustness of the front-end visual odometry (VO) and cannot represent the interactive targets (\eg humans) of the robot.
The reconstruction of dynamic human object is challenging for the SLAM system.
Because, first, tracking of non-rigid surfaces poses difficulties for the visual odometry of moving cameras. Second, non-rigid surface representation is very expensive for 3D rendering.

In this paper, we focus on the human rich dynamic environments and propose a real-time \textbf{dynamic SLAM} solution with \textbf{ human object recovery}.
%novel framework to recover the dynamic human shapes in a dense RGB-D SLAM scheme. 
%
We firstly follow the human segment removal idea of PoseFusion (PF) \cite{pf} --- adopting learning-based method \cite{bochkovskiy2020yolov4} detect and remove human objects for precise static background reconstruction. 
We then recover them using SMPL \cite{smpl} human models for robot interaction.  
Finally, the recovered meshes, human motion trajectories, and camera trajectory are presented within the reconstructed static background maps.   
%In this work, we build on the human pose detection method of PoseFusion~\cite{pf}, and extend to the recovery of human shape by involving the SMPL \cite{smpl} human model. 
%
%To the best of our knowledge, this is the first real-time \textbf{dynamic SLAM} solution with \textbf{ human object recovery}. 
Our contributions are:
\begin{itemize}
\item A fast learning-based method to track moving humans and to filter the background's residual human bodies (around 26 \emph{fps} without online mesh rendering).
\item An improved loop closure in long distance/large area dynamic human environments.
\item A fast human pose and shape recovering pipeline to stack the dynamic human meshes with their moving trajectories.
\end{itemize}
%The source code of this paper will be open-sourced upon acceptance. %at: \url{https://github.com/UoE-AIRS/posefusion2}

\begin{figure}[tbp]
\begin{minipage}[t]{\linewidth}   
\includegraphics[width=\columnwidth]{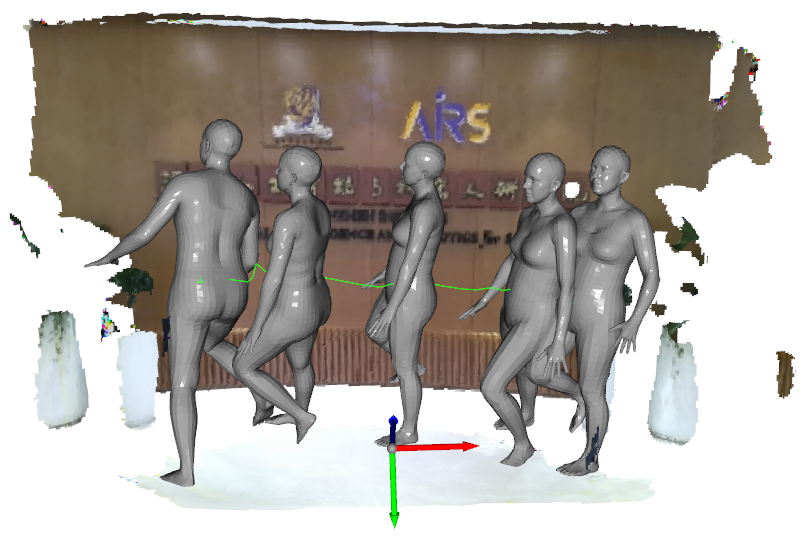}
  \end{minipage}% 
    \hfill
  \begin{minipage}[t]{0.18\linewidth} 
    \includegraphics[width=\columnwidth]{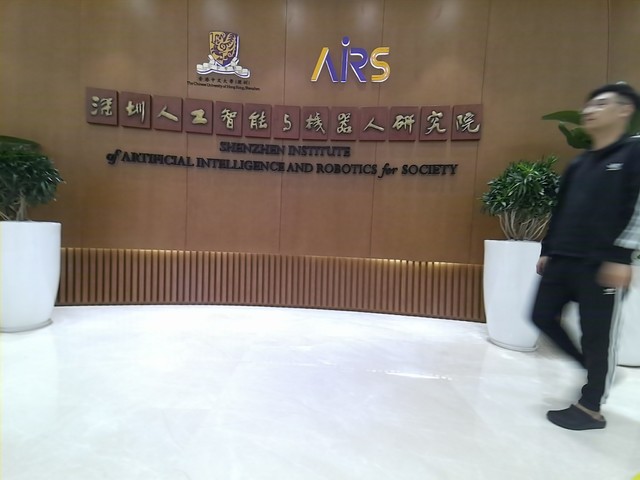} 
 \noindent \caption*{Frames: 62}
  \end{minipage} 
    \hfill
  \begin{minipage}[t]{0.18\linewidth} 
    \includegraphics[width=\columnwidth]{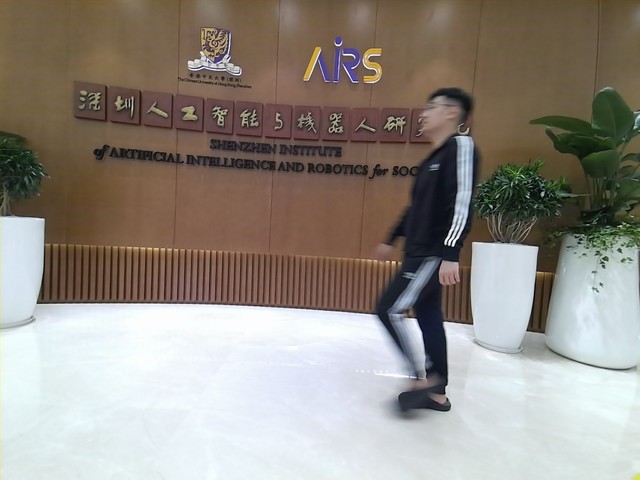}
  \noindent \caption*{73}
  \end{minipage} 
  \hfill
  \begin{minipage}[t]{0.18\linewidth} 
  \includegraphics[width=\columnwidth]{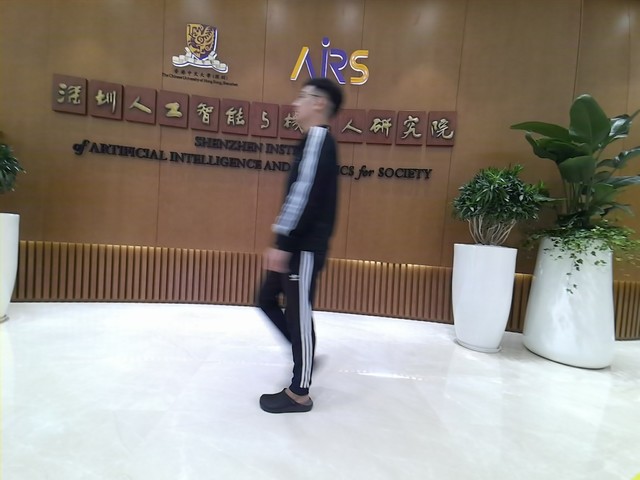} 
 \noindent \caption*{82}
  \end{minipage} 
  \hfill
  \begin{minipage}[t]{0.18\linewidth} 
  \includegraphics[width=\columnwidth]{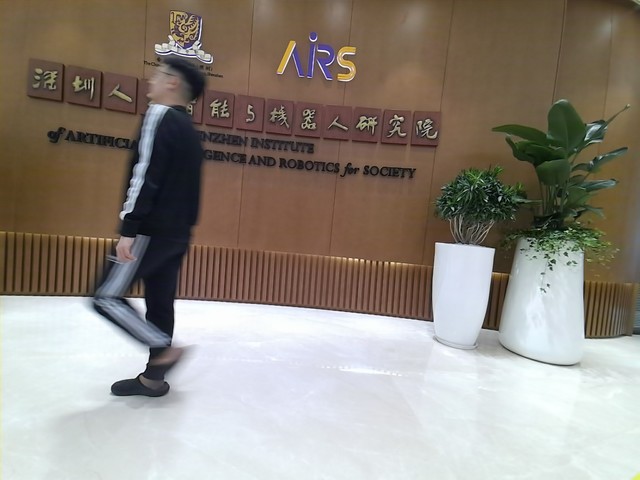} 
 \noindent \caption*{91}
  \end{minipage}% 
  \hfill
  \begin{minipage}[t]{0.18\linewidth} 
  \includegraphics[width=\columnwidth]{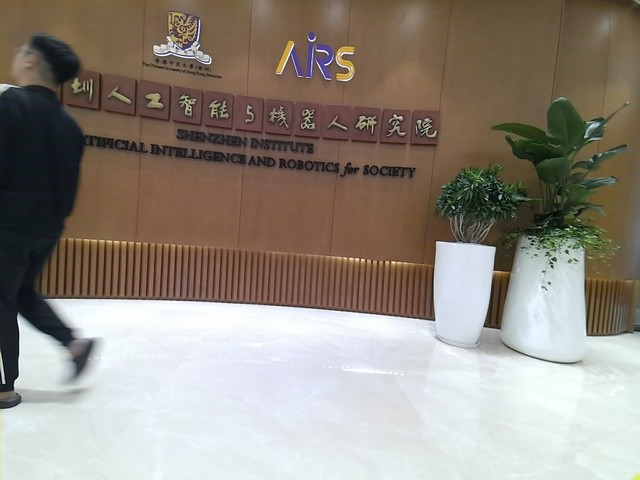} 
 \noindent \caption*{100}
  \end{minipage} 
\caption{Dynamic Scene Reconstruction. Our proposed method performed dynamic human object removal,  tracking and recovery and static environment mapping in 26 \emph{fps}. The green line is the estimated human moving trajectory.}
\vspace{-0.6 cm}
\label{Fig1}
\end{figure}

%%%%%%%%%%%%%%%%%%%%%%%%%%%%%%%%%%%%%%%%%%%%%%%%%%%%%%%%%%%%%%%%%%%%%%%%%%%%%
%%%%%%%%%%%%%%%%%%%%%%%%%%%%%%%%%%%%%%%%%%%%%%%%%%%%%%%%%%%%%%%%%%%%%%%%%%%%%%%%

\section{Related Works}
\begin{figure*}[tbp] 
	\centering
	\includegraphics[width=\linewidth,scale=1.00]{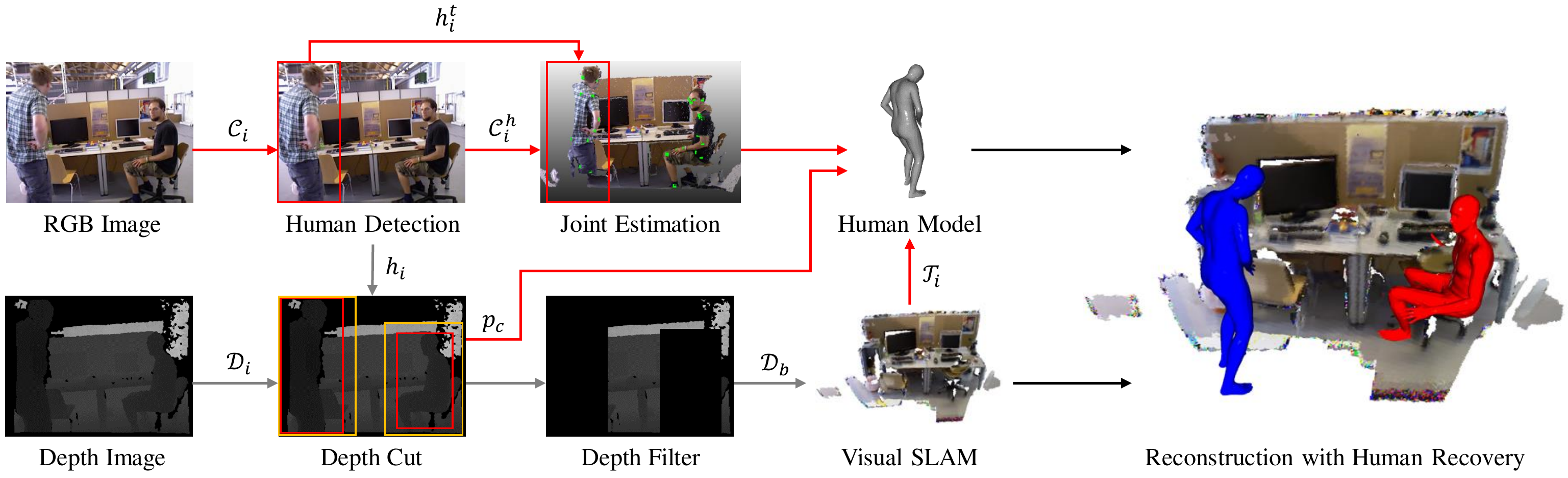}
	\caption{Approach Flowchart. 
	%We propose a {full automatic real-time} dynamic human object recovery and real environment reconstruction approach. 
	 We adopt a fast Dual Bounding Box (the yellow and red boxes) object detector to separate humans objects from backgrounds. The segmented human shapes are recovered to human models and finally represented together with reconstructed backgrounds.
%To improve the robustness of global loop closure performance, the BundleFusion \cite{bf} is adopted as our baseline approach.  
}
	\vspace{-0.5cm }
\label{Framework}
\end{figure*}

\subsection{\textbf{Dynamic SLAM solutions}}
Most of the existing dynamic SLAM solutions try to deal with the dynamic environment problem by finding and removing dynamic objects. Based on their object recognition approaches, 
we divide the current state-of-the-art into motion segmentation-based and object detection-based methods. 

\textbf{Object detection-based dynamic SLAM methods}
usually utilize advanced deep learning-based object detectors to remove dynamic objects, and then enable the classical static SLAM frameworks in the dynamic environments.
Zhang \etal \cite{pf} combined Openpose \cite{op} skeleton estimator into ElasticFusion (EF) \cite{ef} to remove dynamic human body point clouds in dense background reconstruction.
Bescos \etal \cite{DynaSLAM} proposed DynaSLAM which applied Mask-RCNN \cite{mr-cnn} to detect human objects in RGB images and adopted ORB-SLAM2 \cite{orb2} for camera tracking.
The Mask-RCNN outputs accurate human silhouettes,
%to the DynaSLAM visual odometery
but it takes abound 300 ms per frame.  

\textbf{Motion segmentation based approaches} attempted to find dynamic pixels or point clouds rather than recognizing moving objects. 
Scona \etal proposed StaticFusion \cite{sf} that combined scene flow computation with EF's Visual VO to achieve real-time static background reconstruction in a small-sized room.
%T. Zhang \etal proposed FlowFusion \cite{ff}, which extended SF by involving optical flow residuals to enhance the moving segmentation performance.
Judd \etal provided a multi-object motion segmentation method in \cite{mvo}, which applied sparse feature points alignment to separate and track multiple rigid objects. 
%non-rigid paragraph need updates

\subsection{\textbf{Dynamic Human Reconstruction}}
%Static Human shape can be easily modeled through 3D scanning, structure from motion technology, or multi-view geometry approach.
%However, dynamic human shape modeling is still a challenging problem. 
The general solution for dynamic object reconstruction in dynamic environments is to omit the rigid backgrounds and explicitly model the scene's non-rigid components.
Traditional methods use the pre-scanned template and match the template to live RGB or RGB-D streams, a recent work based on this idea is \eg livecap \cite{livecap}. They developed the deep learning framework of template-based human motion tracking and obtained real-time performance.
Another is the volumetric approach. Newcombe \etal proposed the first template-free  on-the-fly framework in \cite{df} for non-rigid surfaces reconstruction.
More recently, SurfelWarp~\cite{surfelwarp} extended this framework and cut down the memory and computation cost by employing Surfel rather than a 3D volume representation.

To deal with the noise and reduce online computation, another group of methods fit learned human models, \eg SMPLify \cite{smpl} and SMPL-X \cite{smplx}, to sparse \textbf{Pose Estimation} results for shape recovery.
Kanazawa \etal proposed Human Mesh Recovery (HMR) \cite{hmr} which developed a Generative Adversarial Network to recover 3D SMPL human shapes from the estimated 2D joint key-points.   
Based on HMR,  Muhammed \etal developed an efficient Video Inference for Human shape Estimation (VIBE) \cite{vibe}.

\subsection{\textbf{Human Shape Recovery within SLAM}}
%smplSLAM
% be careful of this saying:
In terms of \textbf{Human Shape Recovery in SLAM} applications, there are not many exemplars yet that work robustly.
The first attempt was \cite{psi}. The authors take dense RGB-D SLAM reconstruction results and SMPL-X human model as input to generate \textbf{Offline Static data}, which contain expressive 3D human bodies that naturally interact with the 3D environment.
Then, Rosinol \etal 
 %The most different point between \cite{sb} and ours is that, 
proposed a human node concept in dynamic SLAM scene and 
a graph CNN approach to regress SMPL vertices and track
the human torsos in \cite{sb}.
%provided a graphics dataset that contains human meshes in \cite{sb}.
%Notwithstanding, they only inserted some SMPL human meshes generated through virtual animation. 
%
More recent, Li \etal proposed SplitFusion (SpF) \cite{spf} which, in parallel,  reconstructs rigid backgrounds and non-rigid human objects. 

Overall, dense key-point tracking based non-rigid dynamic human reconstructing methods such as \cite{spf} \cite{df} \cite{surfelwarp} are able to show more surface details \eg hairs, clothes folds and emotions. However, it's hard to improve noise robustness and online efficiency.
Model-based human shape estimation approaches, such as \cite{smplx} \cite{vibe} \cite{hmr} can efficiently output barebone 3D human meshes with the help of fast learning-based 2D human pose estimation tools. This scheme is promising for real-time systems and the basic human mesh is good enough for Human Robot Interactive (HRI) applications. However, these model based methods require rich body joint movement input to ensure accurate whole body shape recovery. This requirement is usually hard to satisfy in SLAM systems where human target moves in close proximity to the mobile robots.

%%%%%%%%%%%%%%%%%%%%%%%%%%%%%%%%%%%%%%%%%%%%%%%%%%%%%%%%%%%%%%%%%%%%%%%%%%%%%%%%
\section{SYSTEM FRAMEWORK AND METHODS}
%%%%%%%%%%%%%%%%%%%%%%%%%%%%%%%%%%%%

%In our study, we develop the human object detection and recovery approach for the real dynamic human scenes without any priors.
In this study, we present a fully automatic system which simultaneously reconstructs static background and recover 3D dynamic human objects in the environments. 
To improve the loop closure performance in dynamic environments, we integrates BundleFusion \cite{bf} as the baseline method.
Different from SpF, as a trade-off between real-time performance and human mesh accuracy, we implement human tracking in feature-based SLAM and reconstruct dynamic 3D human silhouette bodies using the model-based method.
The method pipeline is shown in Fig. \ref{Framework}. The various components of the pipeline are detailed in the following section:
1) we adopt a fast Bounding Box (BBox) object detector to separate humans from backgrounds (Section~\ref{sec3.1});
2) we remove and track the moving human objects for static backgrounds reconstruction using a dual BBox method (Section~\ref{sec3.2});
3) finally the separated human segments are fed into 2D pose and 3D shape estimation (Section~\ref{sec3.3}).

\subsection{Fast Human Detection and Outlier Filtering} 
An RGB-D stream is taken as input, where the image pair is denoted as $f_i=( \mathcal{C}_i, \mathcal{D}_i)$, where $\mathcal{C}_i: \Omega \rightarrow \mathbb{R}^3$ and $\mathcal{D}_i: \Omega \rightarrow \mathbb{R}$ stand for the $i$-th color image and depth image, respectively. $\Omega \subset \mathbb{R}^2$ is the image domain. 
Fig. \ref{Framework} shows the system framework, each human body is detected by a learning-based algorithm, and human detection is refined based on geometric relationship. The RGB frame $\mathcal{C}_i$ is first input into YOLOv4 \cite{bochkovskiy2020yolov4} for human detection.
YOLOv4 is a one-stage detection method that can detect 80 types of objects in the RGB image. To speed up object detection as much as possible, we applied an accelerated implementation of YOLOv4 named tkDNN. It speeds up the inference of YOLOv4 while maintaining accuracy.

\begin{algorithm}[tb]
	\renewcommand{\algorithmicrequire}{\textbf{Input:}}
	\renewcommand{\algorithmicensure}{\textbf{Output:}}
	\caption{Human Detection and Outlier Filtering}
	\label{a1}
	\begin{algorithmic}[1]
		\REQUIRE Color image $\mathcal{C}_i$ and Depth image $\mathcal{D}_i$ 
		\ENSURE Human detection $\mathcal{D}_h$ and Filtered background $\mathcal{D}_b$
		\STATE{$h \leftarrow human\_detection(\mathcal{C}_i)$}
		\FOR{each $h_i^t \in h_i$}
		\STATE{$\mathcal{D}_h^t,\mathcal{D}_b \leftarrow scene\_separation(\mathcal{D}_i, h^t_i)$}
		\STATE{$h_{min}^t, h_{max}^t \leftarrow compute\_histogram(\mathcal{D}_h^t)$}
		\ENDFOR
		\STATE{$b_{min}, b_{max} \leftarrow compute\_histogram(\mathcal{D}_i)$}
		%\STATE{$\mathcal{D}_i^w, \mathcal{D}_i^h \leftarrow get\_image\_size(\mathcal{D}_i)$}
		\FOR{each $h_i^t \in h$}
		%\STATE{// Expand $h_t$ by lambda = 1.2 times and control the boundary}
		\STATE{$x_{ul}^t, y_{ul}^t, x_{lr}^t, y_{lr}^t \leftarrow magnify\_range(\lambda * h^t_i)$}
		\FOR{$u \in (^ix_{ul}^t, ^ix_{lr}^t), v \in (^iy_{ul}^t, ^iy_{lr}^t)$}
		\IF {$\mathcal{D}_b(u,v) \notin (b_{min}-0.1, b_{max}+0.1)$}
		\STATE{$\mathcal{D}_b(u,v) \leftarrow 0$}
		\ENDIF
		\IF {$\mathcal{D}_h^t(u,v) \notin (h_{min}^t-0.2, h_{max}^t+0.2)$}
		\STATE{$\mathcal{D}_b(u,v) \leftarrow \mathcal{D}_h^t(u,v)$}
		\STATE{$\mathcal{D}_h^t(u,v) \leftarrow 0$}
		\ENDIF
		\ENDFOR
		\ENDFOR
	\end{algorithmic}
\end{algorithm}

Once the object is detected, the categories, bounding boxes, and confidence of all objects in the scene can be derived.
The bounding box is composed of the coordinate values of the upper left point $(x_{ul}, y_{ul})$ and the lower right point $(x_{lr}, y_{lr})$ in the image plane. We store $N$ bounding boxes containing human bodies with confidence $c>0.5$ as a set $h_i=\{h_i^1,h_i^2,...,h_i^N\}$, where $N$ is the number of humans detected in $\mathcal{C}_i$. We use $h^t_i=(^ix_{ul}^t, ^iy_{ul}^t, ^ix_{lr}^t, ^iy_{lr}^t)$ to represent the $t$-th bounding box in $h_i$.
The detected human body is treated as the non-rigid object in our framework, which is only used for human motion tracking and recovery. The static part will be used for performing SLAM.

However, it is not feasible to remove the range of the human body in the depth image directly based on the detection results in the RGB image. The reasons are as follows:
\begin{itemize}
	\item The object detector may output an incomplete bounding box of human detection.
	\item The depth image and color image are not strictly aligned, which may lead to the human bodies' residual.
	\item The high-speed movement of the human and the depth camera results in the blur of the captured image, and the depth ranging based on ToF will be disturbed.
\end{itemize}

To deal with these problems, we remove the remnant outliers based on geometric features in the depth image. 
Considering that the outliers are near the bounding boxes $h_i$, their depth values must be close to the human body. 
We use histogram statistics to detect abnormal depth values. The data distribution of the human part will differ significantly from the background. 
We first divide $\mathcal{D}_i$ into background $\mathcal{D}_b$ and human parts based on $h_i$. Each human part is denoted by $\mathcal{D}_h^t$.
Then, we can get the histograms of $\mathcal{D}_b$ and each $\mathcal{D}_h^t$.
For the interval with the most occurrences, we use it as the principal component of $\mathcal{D}_b$ and each $\mathcal{D}_h^t$. 
It means that the depth of $\mathcal{D}_b$ is mainly in interval $(b_{min}, b_{max})$, and each $\mathcal{D}_h^t$ is mainly in interval $(h_{min}^t, h_{max}^t)$.
To preserve more background details, we filter the abnormal depth near each $h_i^t$ and recover parts of the background.
More specifically, we extend the filtering range to $\lambda * h_i^t$ to filter boundary depth residuals. $\lambda$ is an experimentally selected constant. 
%that is expanded by 1.2 times.
After the above steps, the filtered background $\mathcal{D}_b$ is used for performing SLAM. The human detection $\mathcal{D}_h$ is input to the next step. The pseudocode for human detection and outlier filtering is presented in Algorithm \ref{a1}. 
\label{sec3.1}

\subsection{Human motion tracking} 
Human motion tracking aims to find the movement trajectory of each human character during SLAM. From the previous stage, we have the human detection $\mathcal{D}_h$ and filtered background $\mathcal{D}_b$. We firstly estimate the camera pose in the world coordinate system $\mathcal{F}_w$. 
The initial position of the camera is taken as the origin of $\mathcal{F}_w$.
Then, we calculate the position of each human in the camera coordinate system $\mathcal{F}_c$. Based on temporal and spatial patterns, we finally get the movement trajectory of each human.

Following the RGB-D fusion framework~\cite{bf}, the camera pose $\mathcal{T}_i \in SE(3)$ is estimated by using an efficient global pose algorithm. 
$\mathcal{T}_i$ denotes a $4\times 4$ rigid transformation relative to $\mathcal{F}_w$. 
It is formulated as an optimization problem of sparse features and dense photometric and geometric constraints. The $\mathcal{T}_i$ can be solved from the energy function:
\begin{equation}
\mathcal{T}_i= \arg \min_{\mathcal{T}_i} \{w_{s}E_{s}(\mathcal{T}_i)+w_{d}E_{d}(\mathcal{T}_i)\}
\end{equation}
in which $E_{s}(\mathcal{T}_i)$ is sparse term for coarse alignment, $E_{d}(\mathcal{T}_i)$ is the dense term for refined alignment, and $w_{s}$ and $w_{d}$ are weights corresponding to the sparse and dense term, respectively. This energy function is solved by a GPU-based algorithm \cite{bf}. Our method only estimates the camera pose in the filtered background $\mathcal{D}_b$ so that we can get the robust motion estimation.

Then, the position of each human relative to $\mathcal{F}_c$ is calculated. We use the center point of each $h_t$ as the reference. Instead of directly extracting the depth values of the center point in the depth frame, we calculate the average depth values $d_{m}$ in each $\mathcal{D}_h^t$. The reasoning is that the center point is not necessarily located on the human body when it is not upright, such as in a sitting pose. The 3D center point $p_c$ relative to $\mathcal{F}_c$ is defined in homogeneous coordinates and computed by the pinhole camera model:
\begin{equation}
p_c=(\frac{x_{ul}+x_{lr}-2c_x}{2f_x}d_{m}, \frac{y_{ul}+y_{lr}-2c_y}{2f_y}d_{m}, d_{m}, 1)^T
\end{equation}
where $f_x,f_y,c_x,c_y$ are the intrinsic parameters of the camera. Each $p_c$ relative to $\mathcal{F}_w$ in the $i$-th frame can be calculated as:
\begin{equation}
^ip_c^w = \mathcal{T}_ip_c
\end{equation}

After these steps, we can get the position of each human body during SLAM. However, it cannot generate a continuous trajectory. 
We only get a series of discrete key points; these points do not correspond to each human body. 
The reason is that the $h_i$ calculated at each moment does not contain a unique identity. 
The human body represented by $h_i^t$ and $h_{i-1}^t$ in the last time may be different. 
To address this, we design a heuristic to generate a continuous trajectory for each human character. 
We assume that if the center point of $h_i^t$ and $h_{i-1}^t$ are closest, they represent the same human.

% \begin{figure}[t]
%     \begin{minipage}[\columnwidth]{1\linewidth}
%     \centering
%     \includegraphics[width=\columnwidth]{fig/p1.png}
%       \caption*{(a) First Human Reconstruction}
%     \end{minipage}
%     \begin{minipage}[\columnwidth]{1\linewidth}
%     \centering
%     \includegraphics[width=\columnwidth]{fig/p2.png}
%       \caption*{(b) Second Human Reconstruction}
%     \end{minipage}
%       \caption{3D Reconstruction of the detected human objects in TUM fr3/walking\_xyz sequence. Two people were reconstructed separately.}
%     \label{fig:human}
% \end{figure}
\label{sec3.2}

\subsection{Human Shape Recovery in Static Reconstruction}
\label{sec3.3}
We use a model-based method to recover the human body. Following the human recovery framework VIBE~\cite{vibe}, we use the SMPL model to express the human shape and pose. 
The 2D human joint is estimated in the range of each $h_i^t$ based on Openpose \cite{op}. 
Each human joint is projected to 3D planes to find the orientation of the human body. 
Then, the SMPL model's parameters are estimated based on pose prior and shape prior. After the above steps, each human's model can be recovered. 
However, the recovered model does not yet contain spatial information. To recover the human model in static reconstruction, we must find their spatial relationship, which we represent by a transformation matrix $\mathcal{T}^w_h$. 

The human's base coordinate system $\mathcal{F}_h$ is located in their waist. It has the same x-axis orientation as $\mathcal{F}_w$, but the y-axis and z-axis are reversed. So we firstly rotate $\mathcal{F}_h$ by 180 degrees about the current x-axis to make $\mathcal{F}_h$ and $\mathcal{F}_w$ the same direction. The human model output by VIBE is inferenced in $\mathcal{F}_c$. To get the posture of the human model under $\mathcal{F}_w$, we then rotate it based on camera pose's rotation part $\mathcal{R}_i \in SO(3)$. Finally, we displace $\mathcal{F}_h$ according to human's 3D center point $^ip_c^w$ under $\mathcal{F}_w$. The final transformation is computed as:
\begin{equation}
\mathcal{T}^w_h = 
\begin{bmatrix} 
\mathcal{R}_i\mathcal{R}_{x,\theta} & ^ip_c^w \\ 
\textbf{0} & 1 
\end{bmatrix}
\end{equation}
where $\mathcal{R}_{x,\theta}$ represents a rotation of angle $\theta$ about the current x-axis and $\theta=180$. After we have inferred $\mathcal{T}^w_h$ and the human model, it can be recovered in the static reconstruction.

\section{Experiments Results and Evaluations}
We evaluate the proposed approach under three metrics: \textbf{visual odometry, scene mapping} and \textbf{time efficiency}. To ensure the scalability of the method to different application scenarios, we test it on both benchmark dynamic SLAM data sets and real world environments. 
The public data sets used include:
\subsubsection{TUM RGB-D SLAM data set \cite{tum-dataset}} A benchmark containing RGB-D and ground-truth data for the evaluation of VO and visual SLAM systems -- the \emph{fr3} sequences are iconic and one of the earliest dynamic SLAM scene standards, complete with  online evaluation tools \cite{url:tum-tools}.

\subsubsection{HRPSlam humanoids robot dynamic SLAM dataset \cite{hrpslam}} 
The first humanoid robot dynamic SLAM data set, HRPSlam2  (23543 frames, 879.2 seconds) scene is very challenging, as it involves several cases of robot's falling and re-initializing.
These sudden motions caused by the robot falling and the discontinuity of visual information predicated higher \textbf{camera relocation} and \textbf{global loop closure} requirements for SLAM approaches. 

In addition, the real-world AIRS indoor dynamic scenes, shown in Fig.~\ref{Fig1} and \ref{fig:airs-trj}, are captured with an Azure Kinect sensor.  
%Note that the AIRS scene camera tracking trajectories cannot be quantitatively evaluated since there is no Ground Truth (GT) tracking data from motion capture.
All the reported results were obtained on a desktop with Intel Core$^{TM}$  i9-9980XE CPU @ 3.00 GHz $\times$ 36, 128 GB System memory and Four GeForce RTX 2080 Ti GPUs.
%%%%%%%%%%%%%%%%%%%%%%%%%%%%%%%%%%%%%%%%%%%%%%%%%%%%%%%%%%%%%%%%%%%%%%%%%%%%%%%%%%%%%%%%%%%%%%%%%%%%%%%%%%%%%%%%%%%%%%%%%%%%%%%%%%%%%%
\begin{figure*}[htb] 
	\centering
	\includegraphics[width=\linewidth]{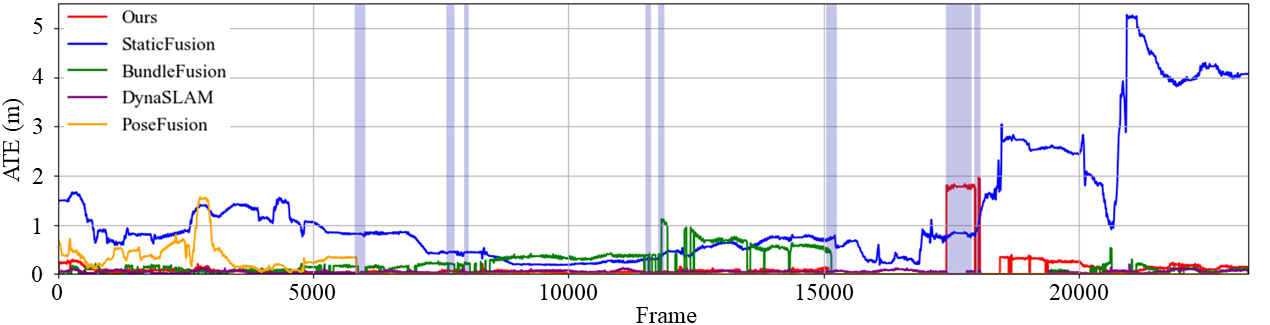}
	\caption{
	Absolute Trajectory Errors (m) comparison in HRPSlam2 sequence.
	%with our proposed method (red), DynaSLAM (purple), PoseFusion (yellow), StaticFusion (blue) and BundleFusion (green).
	The shaded areas indicate the five times camera was violently disturbed from robot falling.
%	Offline method DynaSLAM gets the smallest ATE.
	Our method achieved the smallest errors amonst the online SLAM methods and succeeded in camera relocation under difficult fall scenarios.
    %PF curve stopped at Frame 5480 because it cannot re-locate the camera pose after the fall. The BF green curve looks small because when it failed to estimate the camera pose, it remains zero until a loop is detected. Note that the BF camera tracking trajectory jumps several times in Fig.~\ref{fig:hrp-LoopPerformance} 
    %Note that ATE trajectories started from a non-zero offset since the evaluation tool \cite{url:tum-tools} tried to align the whole global trajectories to the GTs.
	}
\label{HRPSlam_ATE}
\vspace{-0.5cm}
\end{figure*}

\begin{table}[htb]
\centering  
\caption{Absolute Trajectory Error RMSE (cm)}
\label{T:ate}  
\begin{tabular}{p{1.6cm}m{1.4cm}<{\centering}m{0.7cm}m{0.7cm}m{0.7cm}m{0.7cm}}
\hline\noalign{\smallskip}
\textbf{Sequence}  & DynaSLAM & SF & PF & BF & Ours\\
\noalign{\smallskip}\hline\noalign{\smallskip}
      fr1/xyz & 1.0 & 1.4  &1.9 & 2.0 & 2.0 \\
      fr1/desk2 & 2.2 & 5.2 & 4.0 & 7.7 & 7.7 \\
\noalign{\smallskip}\hline\noalign{\smallskip}
fr3/walking\_xyz& 1.7  & 9.2  & 4.8 & Fail& 5.1\\
HRPSlam2.1 & \textbf{4.2} & 51.4  & 31 & 7.5 & 7.1\\
\textbf{HRPSlam2} & \textbf{5.4} & 174  & Fail & 34.5 & 10.8\\ 
\noalign{\smallskip}\hline\noalign{\smallskip}
\end{tabular}
\vspace{-0.5cm}
\end{table}

\begin{figure}[tb]
	\centering
	\includegraphics[width=8.5cm]{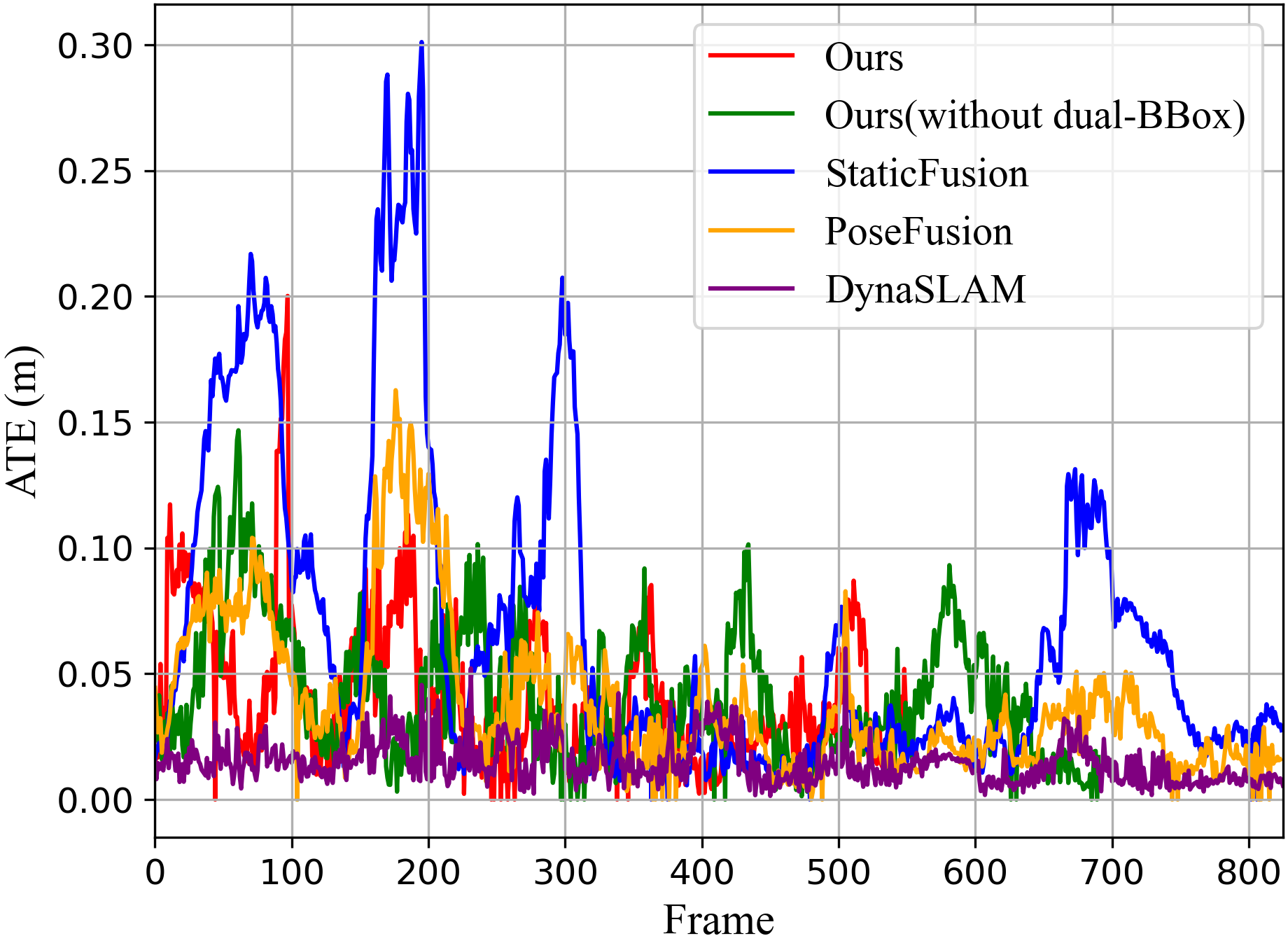}
    \caption{The ATE of SOTA SLAM methods in the TUM $fr3$ dynamic sequence.
    Our method ATE was close to PF and better than SF. %$9.2~cm$. % DynaSLAM achieved the best camera tracking result $1.7~cm$.
    %since it spent much time on accurate human mask segmentation. 
    %BF result was not shown since it failed in this sequence. 
   % The green curve indicated that without the proposed dual-BBox filter, the VO robustness obviously decreased.
    %$6.35$ cm/s for RPE RSME.
    }
    \label{Fig:tumATE}
    \vspace{-0.5cm}
\end{figure}
\subsection{Visual Odometry Evaluations}
The SLAM solution's VO is usually evaluated by the camera tracking  Absolute Trajectory Errors (ATE). We compute ATE's root-mean-square deviation (RSME) via the tools in \cite{url:tum-tools}. 
We compared the proposed method with state-of-the-art (SOTA) dynamic environment reconstruction methods:
%We compare with state-of-the-art dynamic SLAM solutions, such as 
DynaSLAM, StaticFusion (SF), PoseFusion (PF) and the baseline approach BundleFusion (BF).  

Table~\ref{T:ate} shows the VO performances on TUM data set sequences (the sequences start from $fr$) and HRPSlam \cite{hrpslam} RGB-D data sets. 
%To compute ATE as defined in TUM data sets \cite{tum-dataset}, 
%we firstly align the estimated camera trajectory to the ground truth using the least-square method and then compute the absolute errors between aligned poses at each timestamp -- to do this, we exploit the tools provided in \cite{url:tum-tools}.
The first two $fr1$ sequences are static environments. All the methods achieve very small errors, which indicate that all methods work equally well in static environments. 
Note that ours and BF have the same error values in static scenes since we adopted BF as the basic method.
%Our method performed as same as the BF in the static scenes. 
The lower three rows are dynamic sequences. In TUM dynamic sequence \emph{fr3/walking\_xyz},
our proposed method achieves $5.1~cm$ ATE, which is better than SF's $9.2~cm$ (see the red curve in Fig.~\ref{Fig:tumATE}). 
Without the Dual-BBox algorithm, the ATE degrades to $5.9~cm$. 
These results indicate that the proposed Dual-BBox strategy (we set $\lambda = 1.2 $ in TUM and HRPSlam experiments, it costs $3.5~ms$ per frame) efficiently improves the VO performance to the SOTA level.
PF spent more than $500~ms$ on human 3D point cloud segmentation to obtain $4.8~cm$ ATE. 
For the other SOTA methods, DynaSLAM achieved the best camera tracking result  of $1.7~cm$ since it spends much time on accurate human mask segmentation. 
BF result is not shown since it failed in this sequence. The reason is that BF divides front-end VO and back-end global loop closure into two pipelines run on two separate GPUs. This framework design greatly improves global mapping performance in large room size scenes.  

%%%%%%%%%%%%%%%%%%%%%%%%%%%%%%%%%%%%%%%%%%%%%%%%
%in HRPSlam dataset
%DynaSLAM obtained the best accuracy as an offline method. 
The plots in Fig.~\ref{HRPSlam_ATE} shows the ATE values along with the frame IDs in HRPSlam2 whole sequence. The curves of different colors stand for: Our method (red), DynaSLAM (purple), StaticFusion (blue) and BundleFusion (green). Note that the ATE trajectories started from a non-zero offset since the evaluation tool \cite{url:tum-tools} tried to align the whole global trajectories to the GTs.
HRPSlam2.1 is the sequence before the robot first fall of HRPSlam2.
All five methods are able to track the camera pose in this sequence, but in the following sequence, the method without robust camera repositioning fails.
In this difficult scene, as our method applied advanced dynamic object detection and removal technique, VO errors are competitive to the other SOTA online methods. 
It achieved $10.8~cm$ ATE which is better than BF ($34.5~cm$) and SF ($174~cm$), as evidenced in the reconstructed maps (Fig.~\ref{fig:hrp-LoopPerformance}). 
Offline approach DynaSLAM obtained a smaller RMSE error of $5.4~cm$ with the help of Mask-RCNN's careful human silhouette segmentation. 
The PF camera tracking comprehensively fails after the humanoids falling at frame 5480 and it cannot re-locate the camera pose after the fall.  
There are two reasons to this. 
Firstly, PF applied Openpose as human object key-point detector, but it failed to find the human objects without heads (In HRPSlam, the camera is mounted on the robot facing the ground).
Secondly, PF and SF are based on the ElasticFuison \cite{ef} framework.
They emphasize local small area loop closure, but they lack robust global loop detection back-ends.
%%%%%%%%%%%%%%%%%%%%%%%%%%%%%%%%%%%
%%%%%%%%%%%%%%%%%%%%%%%%%%%%%%%%%%%%%%%%%%%%%%%%%%%%%%%%%%%

\label{sec:loop-finding}
%high our loop closing

\begin{figure*}[tbp]
	\centering
	\begingroup
	\setlength{\tabcolsep}{3pt} % Default value: 6pt
	\renewcommand{\arraystretch}{1} % Default value: 1
	\begin{tabular}{cc}
	\includegraphics[height=4cm]{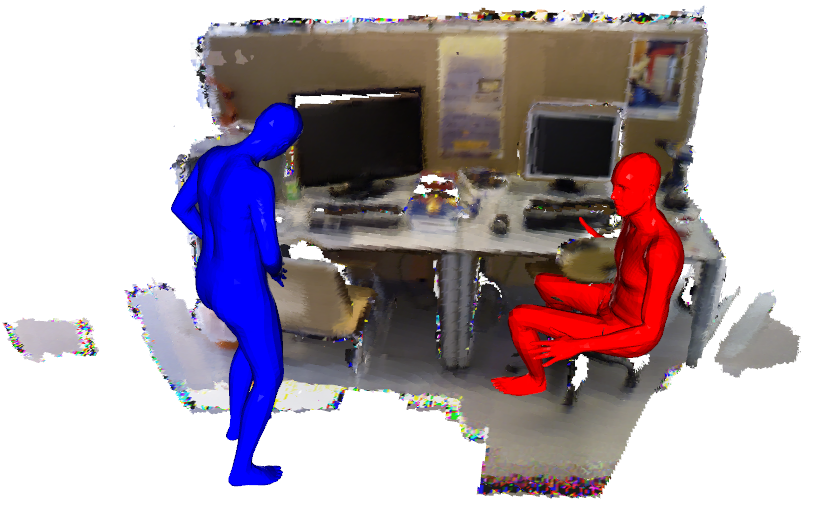} &
	\includegraphics[height=4cm]{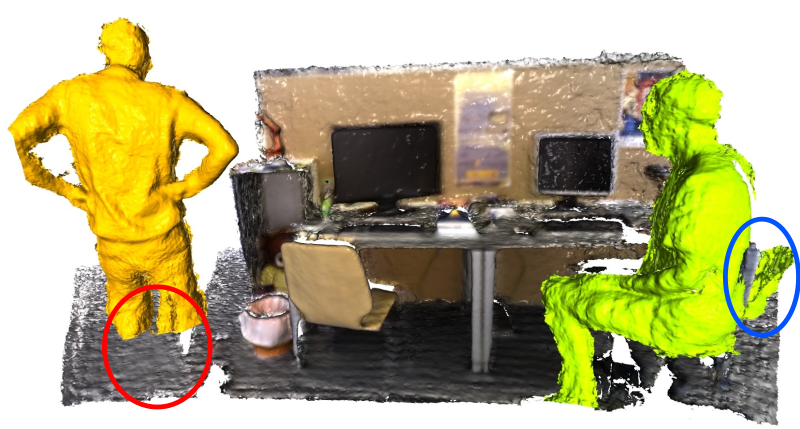} \\
	%Input Image & 
    Ours &
    SplitFusion\\
	\end{tabular}
	\endgroup
	\caption {Reconstruction results on \emph{fr3\_walking\_xyz} sequence of TUM dataset. The input images are as same as Fig.~\ref{Framework}.
    On the left, our method outputs complete and clear human meshes.
    On the right, SpF failed to track the fast subject (see missing feet circled in red). In addition, it cannot distinguish connected objects, \eg in the blue circle, the chair was fused into the green human mesh.
}
	\label{fig:hum-recover}
%	\vspace{-0.5cm }
\end{figure*}

\begin{figure*}[tb]
	\centering
	\begingroup
    \begin{tabular}{m{4cm}<{\centering}m{4cm}<{\centering}m{4cm}<{\centering}m{4cm}<{\centering}}
	\includegraphics[height=3cm]{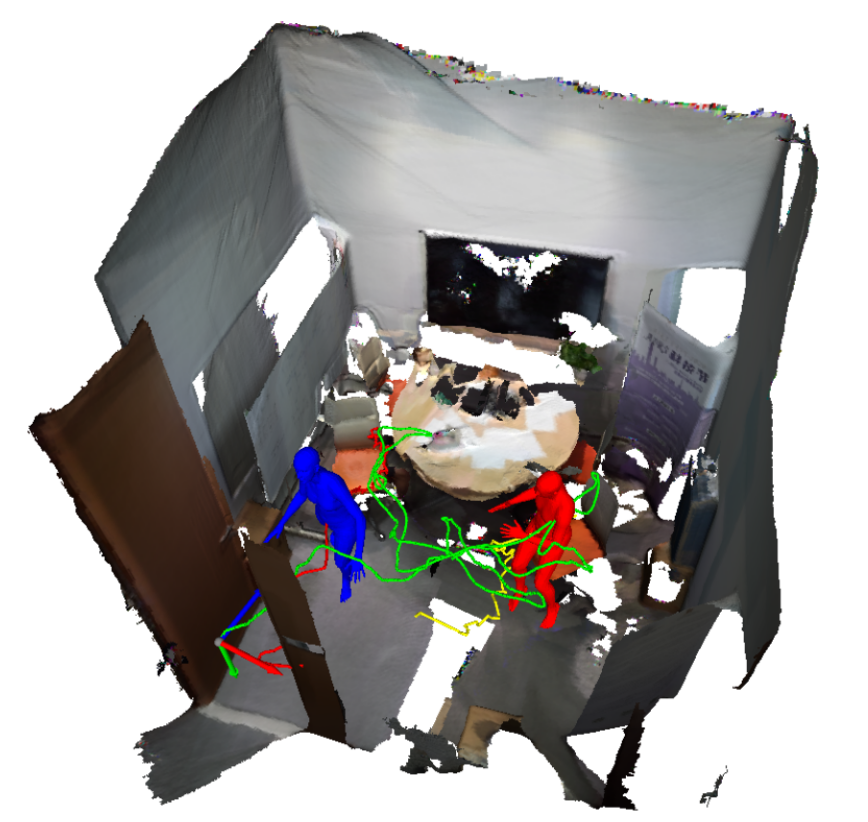} &
    \includegraphics[height=3cm]{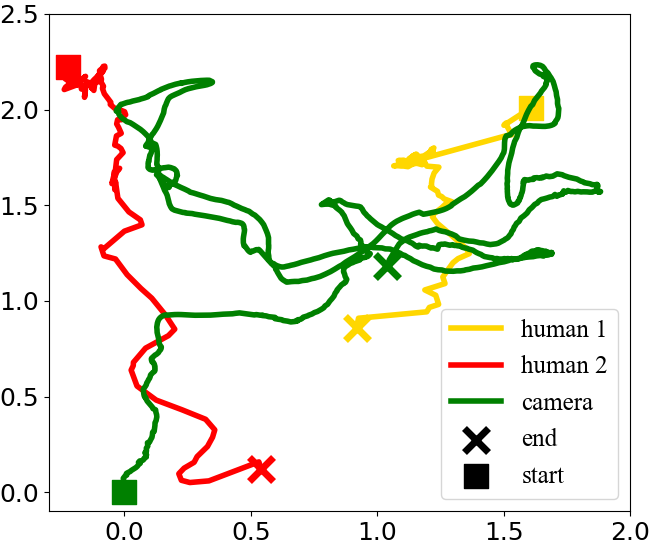}&
     \includegraphics[height=3cm]{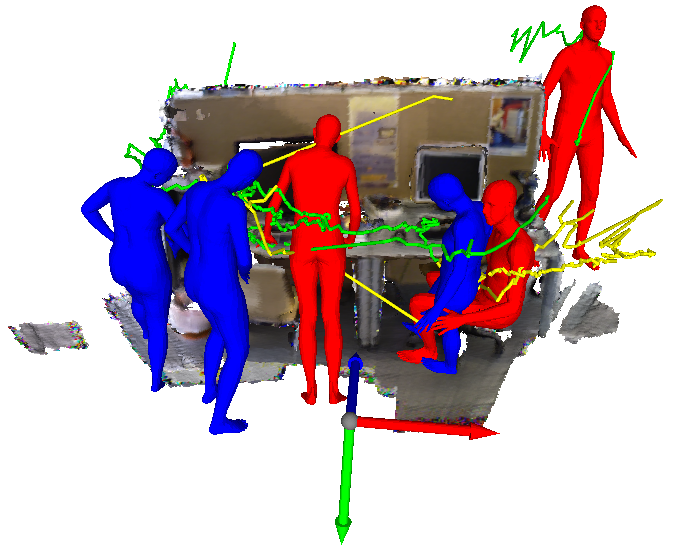}&
     \includegraphics[height=2.8cm]{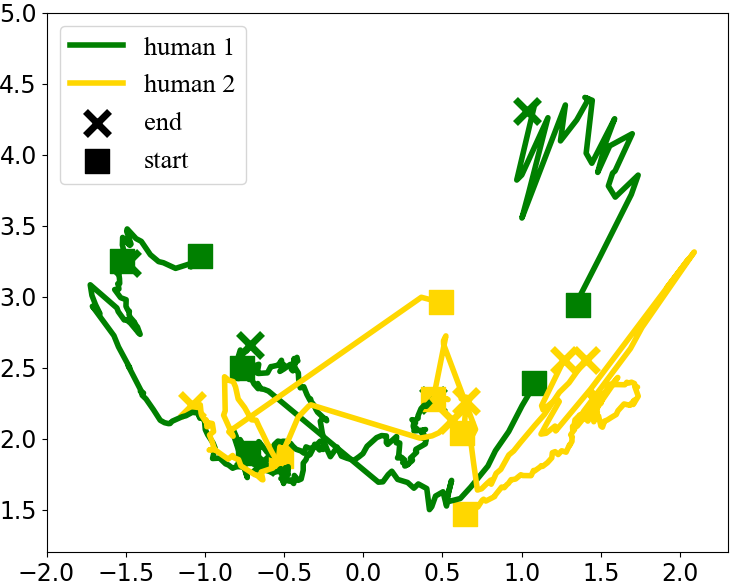}%
     \\
    (a) AIRS scene& 
    (b) AIRS scene trajectories&
	(c) TUM $fr3$ scene &
    (d) TUM $fr3$ trajectories \\
	\end{tabular}
	\endgroup%
	\caption {Scenes reconstruction with human meshes and trajectories. 
	(a), AIRS room scene. 
    (b), the estimated trajectories in (a).
	(c), TUM $fr3$ scene, three recovered meshes for each object were represented to show their motions. (d), human object trajectories of (c).  
	%The camera trajectory is green, the blue human trajectory is in red, and the red human trajectory color is yellow.
    	}
  % \vspace{-0.5 cm}
	\label{fig:airs-trj}
\end{figure*}
We evaluate the scene reconstruction performance of the dense SLAM methods from two aspects. The first is the performance of human object representation. The second is the ability to close the global loop in dynamic scenes.
%Secondly, we evaluate human shape recovery performance.  Since we do not have access to human object pose and shape ground truths, we can only present the recovered human mesh and trajectories in the different scenes. 
\subsection{Human Object Representation Evaluations}
SplitFusion (SpF) and our method has the ability to simultaneously recover human shape and static backgrounds. SpF implements a VO and non-rigid tracking parallel pipeline. Its VO thread is exactly the same as PF, %Thus we only evaluate its human object recovery and time cost. Refer to \cite{spf}, SpF's
thus, its static background reconstruction performance is similar to ours. For the foreground human objects, SpF reconstructs the human mesh via non-rigid tracking, while we replace the human object by the estimated SMPL mesh.
The proposed approach is superior to SpF in three perspectives. 
Firstly, Fig.~\ref{fig:hum-recover} shows the $fr3\_walking\_xyz$ sequence scene reconstructions, from which it can be found that SpF non-rigid tracker can not track fast-moving objects, \eg calves, feet, and hands, while our model-based method represents the complete human shapes, see the red circle area. 
Moreover, SpF cannot distinguish connected objects, for instance, the chair was fused into the green human object. Such mesh representation is hard to recognize as an interactive target.
As a comparison, our method output complete and clear human meshes.
%: Furthermore, Our model-based human shape estimation scheme is more efficient than SplitFusion's graph-based dense key-points tracking scheme.
%
In addition, our method can effectively track moving human targets by BBox center trajectory.
As shown in Fig.~\ref{fig:airs-trj}, the two examples of AIRS and TUM scenes demonstrate that our method can insert a recovered human subject at a location along the estimated trajectory.
%Two examples of AIRS and TUM scenes are shown in , which indicate that our method can inserted back recovered human subject along a position on the estimated trajectories. 
In (a), we inserted the objects to their last seen positions; the 2D estimated object trajectories together camera's were plotted in (b). In the TUM scene (c), we represented three recovered meshes for each object to show their pose changes. Since TUM camera moved in a vertical plane, it was not plotted in (d). 

%To evaluate the scene reconstruction performance, firstly, we compare the qualities of the output maps. 
\subsection{Scene Reconstruction Evaluations}
\begin{figure*}[t]
\newcommand{\hi}{2.8cm}
\centering
\footnotesize
  \begin{tabular}{m{1.3cm}m{3.5cm}<{\centering}m{3.5cm}<{\centering}m{3.5cm}<{\centering}m{3.5cm}<{\centering}}%m{2.7cm}<{\centering}m{2.2cm}<{\centering}}
   % & \multicolumn{5}{c}{rgb images}&  \\
  Frame ID  & 5 & 360 & 770 & 1146\\
   RGB input&
    \includegraphics[height=\hi,valign=m]{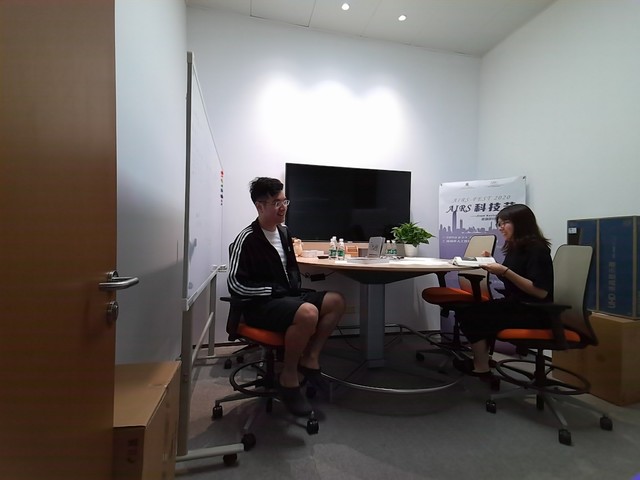} &
    \includegraphics[height=\hi,valign=m]{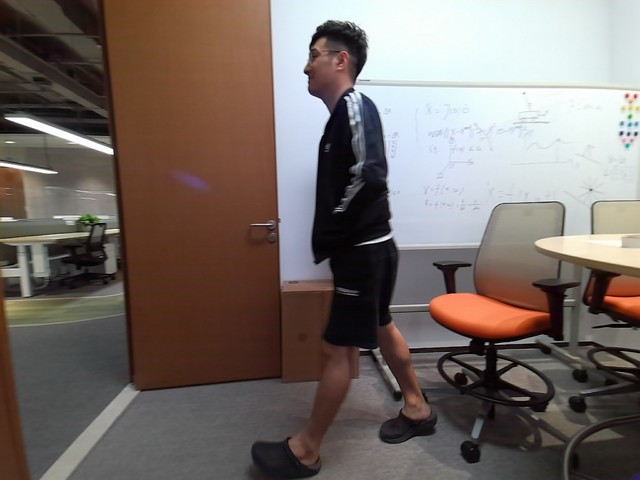} &
    \includegraphics[height=\hi,valign=m]{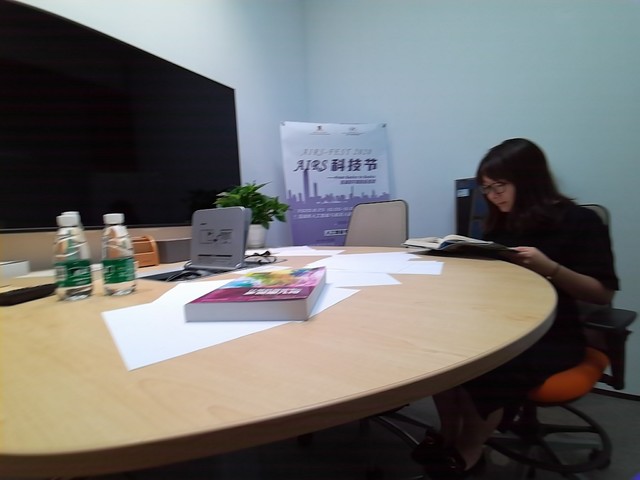} &
    \includegraphics[height=\hi,valign=m]{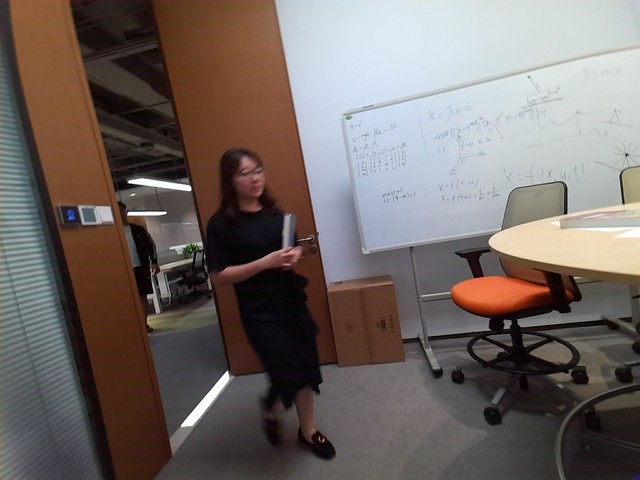} \\
      DynaSLAM &
    \includegraphics[height=\hi,valign=m]{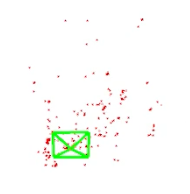}  &
    \includegraphics[height=\hi,valign=m]{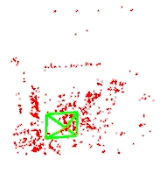}  &
    \includegraphics[height=\hi,valign=m]{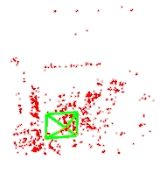} &
    \includegraphics[height=\hi,valign=m]{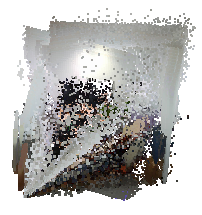}  \\ 
    SF&
    \includegraphics[height=\hi,valign=m]{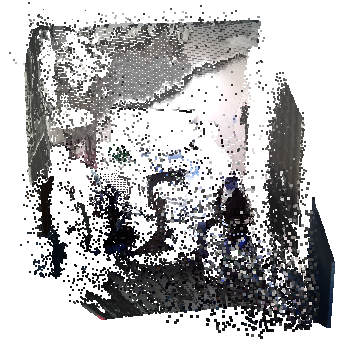}  &
    \includegraphics[height=\hi,valign=m]{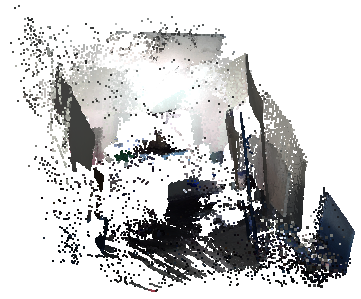}  &
    \includegraphics[height=\hi,valign=m]{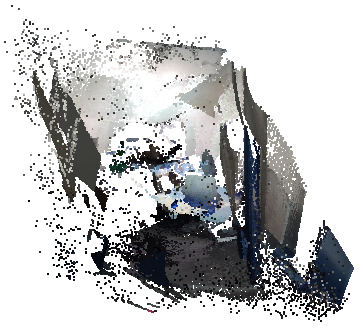} &
    \includegraphics[height=\hi,valign=m]{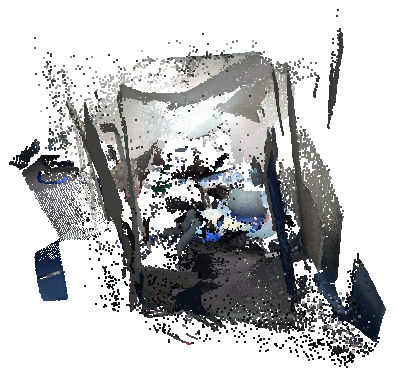} \\
    BF&
    \includegraphics[height=\hi,valign=m]{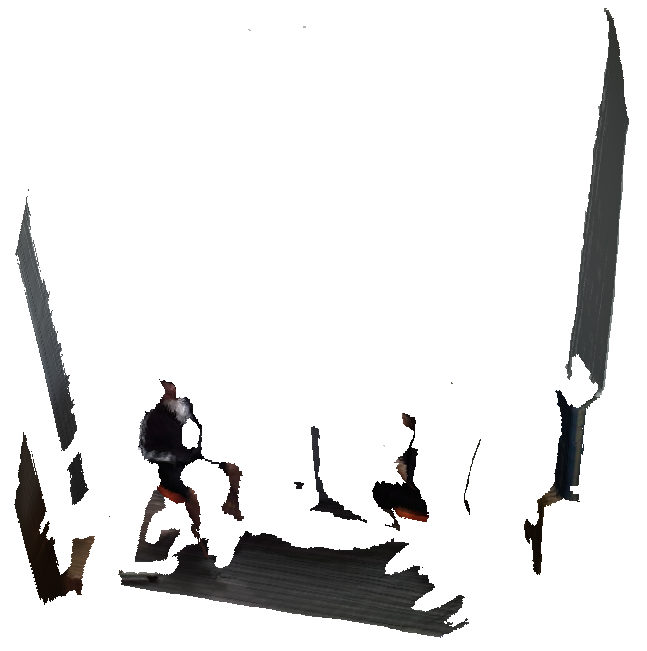}  &
    \includegraphics[height=\hi,valign=m]{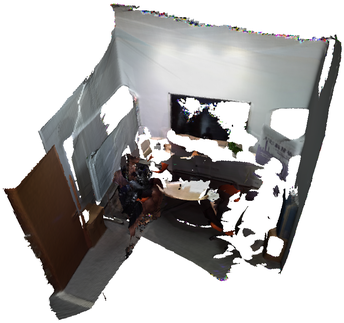}  &
    \includegraphics[height=\hi,valign=m]{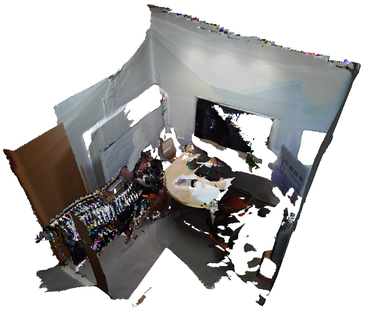} &
    \includegraphics[height=\hi,valign=m]{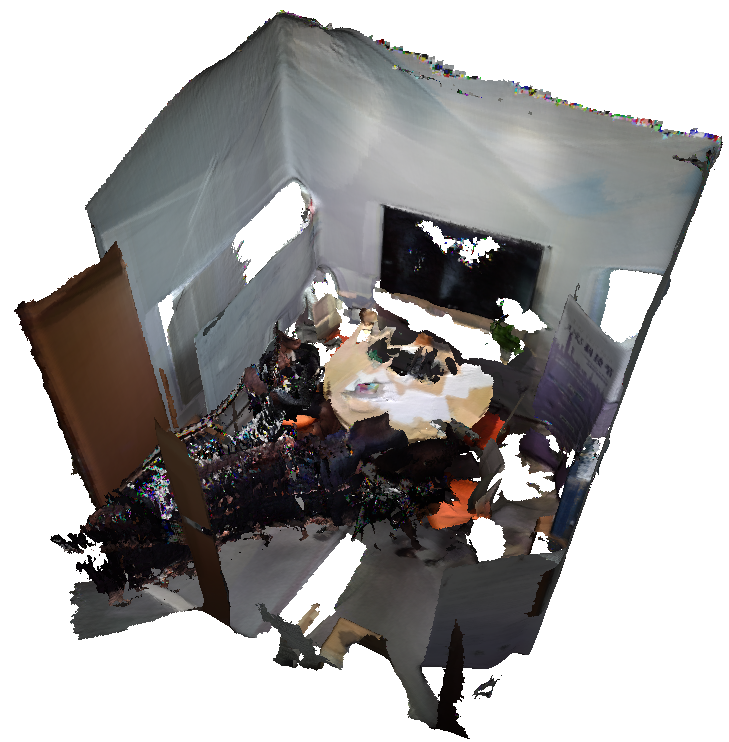}  \\
Ours &
    \includegraphics[height=\hi,valign=m]{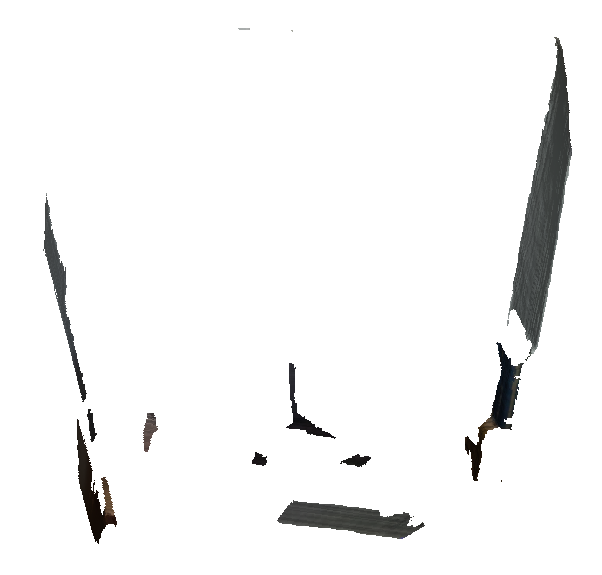} &
    \includegraphics[height=\hi,valign=m]{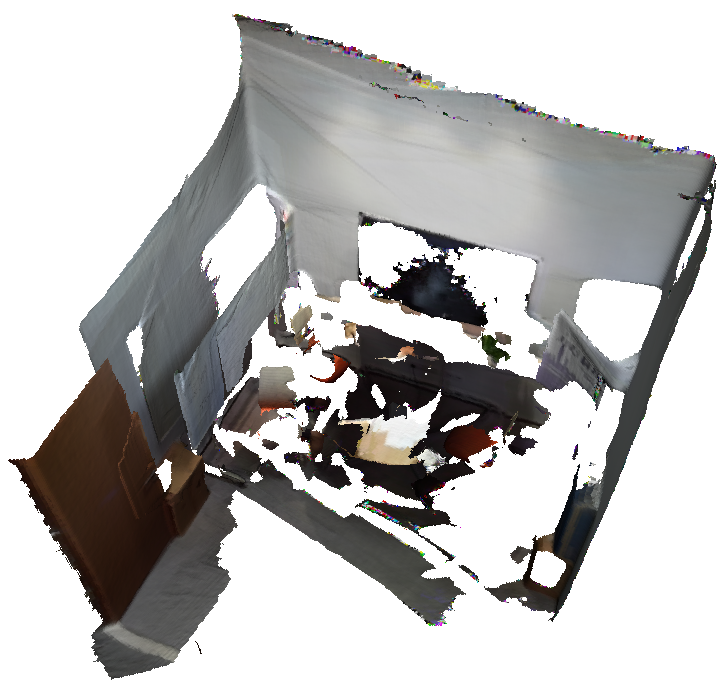} &
    \includegraphics[height=\hi,valign=m]{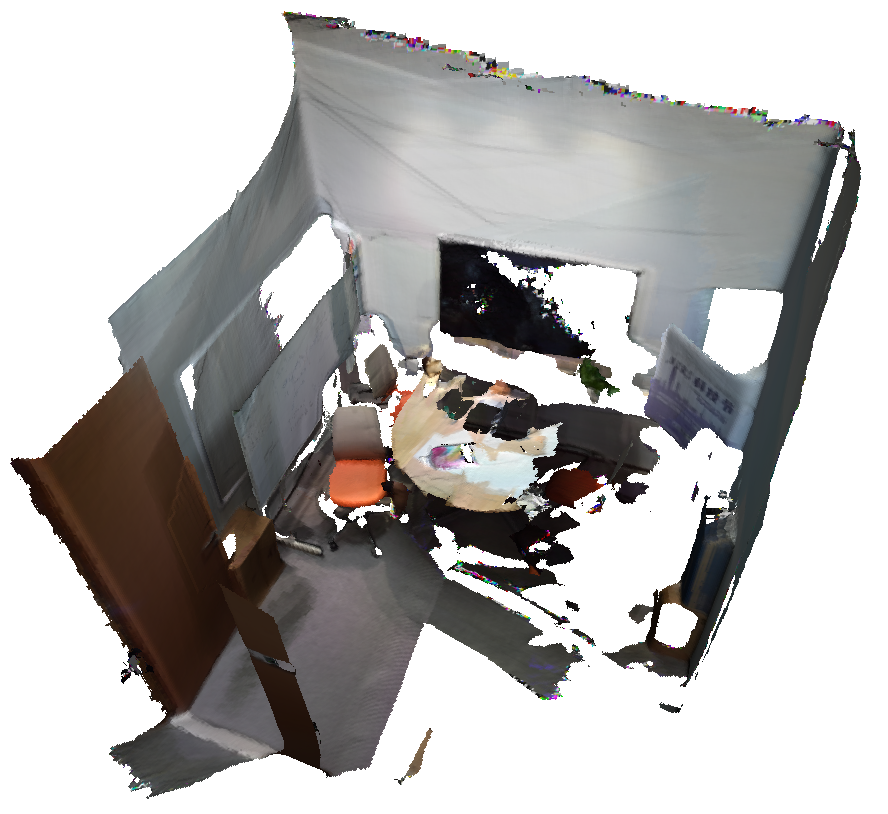} &
    \includegraphics[height=\hi,valign=m]{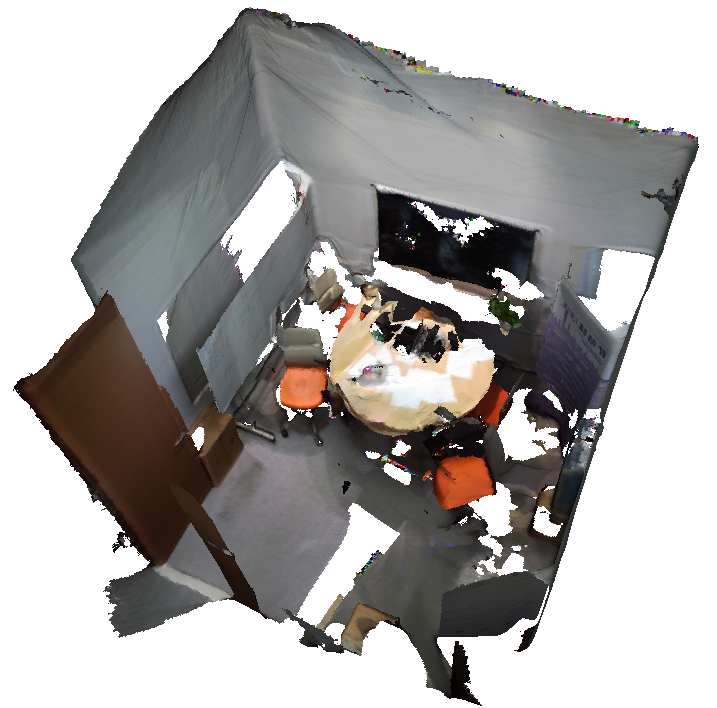} \\
  \end{tabular}
  \caption {AIRS real dynamic scene reconstruction proceedings. As DynaSLAM is an online VO and offline mapping approach, the camera tracking with sparse ORB feature points are shown with frame IDs. For SF, BF, and our proposed method, the resultant mappings are shown. Please refer to text for discussion of the results}
  \label{fig:airsroom-mappings}
 %\vspace{-0.5cm}
\end{figure*}

Our Dynamic environment mapping performance of a real dynamic scene is shown in Fig.~\ref{Fig1}. 
This sequence was captured in AIRS using an Azure Kinect RGB-D sensor. The proposed method performed dynamic human object removal, static environment mapping, and human object tracking and recovery at 21 \emph{fps}, with the green line indicating the recovered human motion trajectory.
Next, Fig.~\ref{fig:airsroom-mappings} captures the results of another AIRS dynamic scene reconstruction with multiple moving subjects. 
For SF, BF, and our proposed method, the mapping proceedings are shown. As DynaSLAM is an online VO and offline mapping approach, in the second row, only the feature-based camera tracking processes are shown. SF kept VO accuracy in the first 300 frames (image 1 to 2, third row), but it did not achieve sufficient loop closure to maintain global mapping. 
The moving objects in the third image result in a big VO drift error for BF, it then reduced these drift by global loop closure after the objects moving out of view. 
Our method removed the moving human objects simultaneously with local mapping. Therefore, we kept VO robustness and further improved the mapping results using BF's key-frame strategy.

\begin{figure*}[h]
\centering
\footnotesize
  \begin{tabular}{m{4cm}<{\centering}|m{4cm}<{\centering}|m{4cm}<{\centering}|m{4cm}<{\centering}}
    \includegraphics[height=2.8cm]{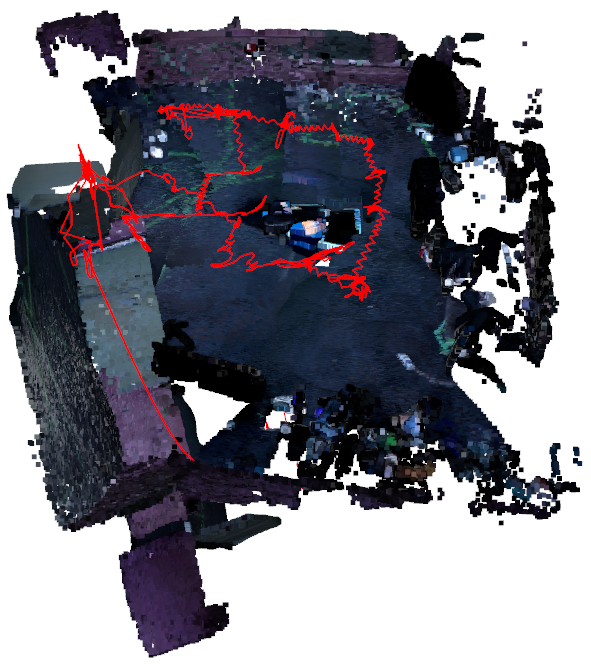} &
    \includegraphics[height=2.8cm]{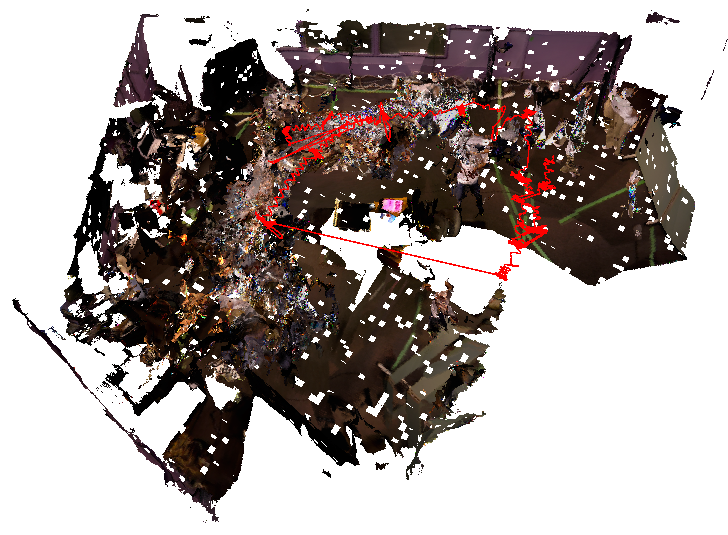} &
    \includegraphics[height=2.4cm]{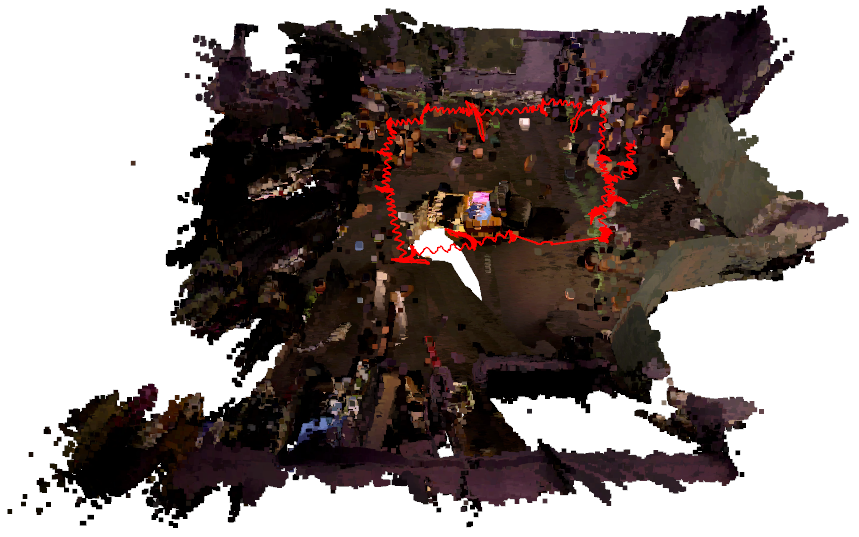} &
    \includegraphics[height=2.8cm]{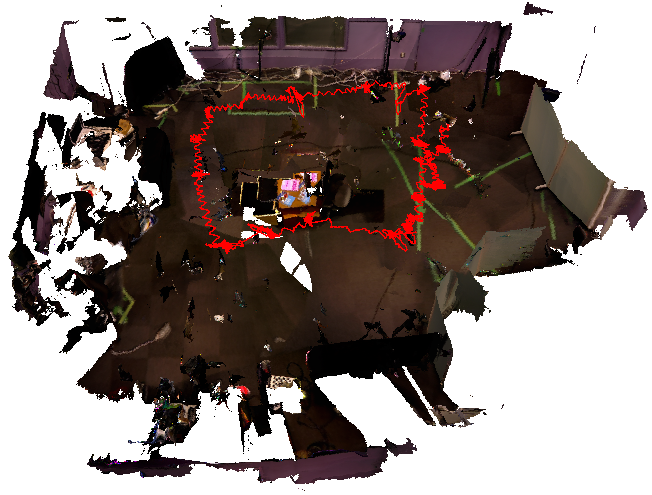} \\
  
    \includegraphics[height=3.6cm]{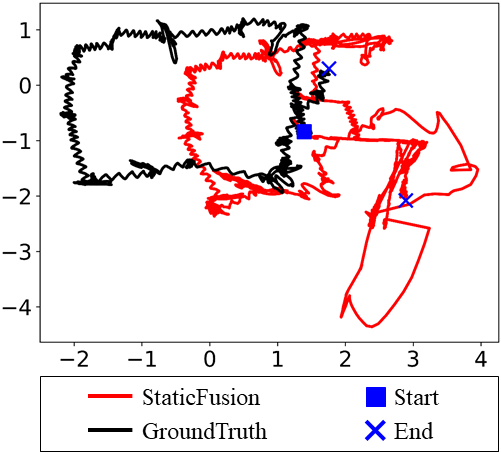}  &
    \includegraphics[height=3.6cm]{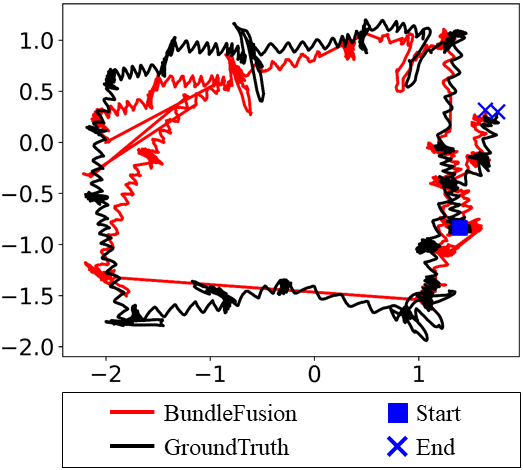}  &
    \includegraphics[height=3.6cm]{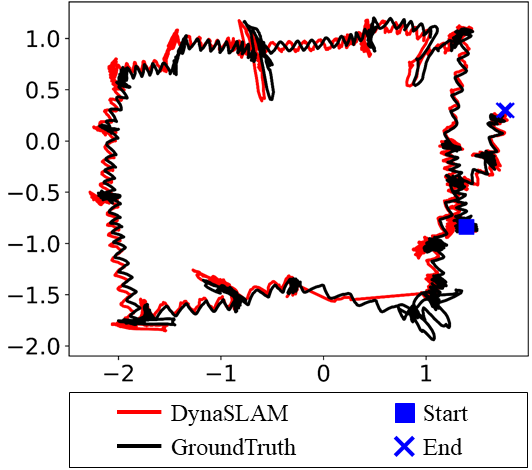} &
    \includegraphics[height=3.6cm]{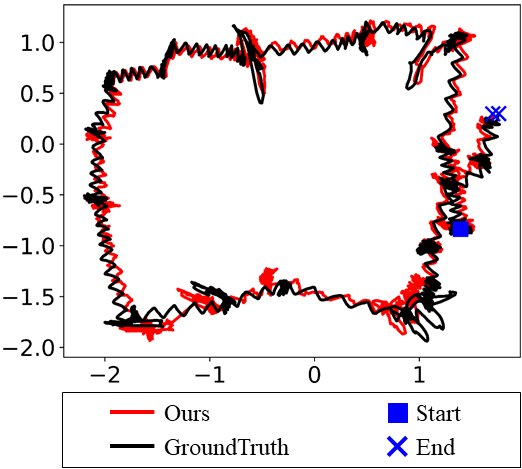}   \\
     %SF & BF & DynaSLAM & Ours \\
  \end{tabular}
  \caption{
Mapping performances in HRPSlam2 sequence that contains difficult humanoid robot falling cases.
From left to right: SF performed wrong loop detection; 
%wrongly aligned the grounds to the wall after the second turn. 
%The reason was that SF emphasized local small area loop closure, but it lack robust global loop detection back-ends. 
BF VO failed in dynamic scene, but it re-located the camera pose after humans moving away.
DynaSLAM also failed in the robot fall near the end;
Our method accomplished excellent re-location and mapping results.
For the details please refer to Section \ref{sec:loop-finding}.
}
\label{fig:hrp-LoopPerformance}
\vspace{-0.6cm}
\end{figure*}

The mapping performance of the HRPSlam2 scene in Fig. \ref{fig:hrp-LoopPerformance} clearly demonstrates the global loop closure capability of these methods in a dynamic environment. 
See the first image, 
The trajectory of SF cannot be aligned with GT. Because it cannot re-locate itself after the camera tracking fails.
In the second image, BF lost robustness in dynamic environments. However, with closed-loop detection, It realigned the camera pose to GT trajectory several times when the obstacle moved away, see the long slash lines.
DynaSLAM remained robust VO performance, except for the failure in the full occlusion case that happened at HRPSlam2 dataset Frame 18570.
To reduce the influence of the dynamic humans on the closed-loop detection,
1) we developed YOLOv4 based dual-BBox human object detector, which enhanced the human detection robustness in such a not full body visible scene. 
%Highlight loop detection performance
2) We selected BF as the basic approach, the advanced key-frame processing and storing strategies contribute to reliable global loop closure. 
With these two improvements, our method accomplished excellent camera re-location and continue the mapping system even under difficult fall scenarios plus full occlusion case.

\begin{table}[tb]
\caption{Computation Speed \emph{fps} of TUM Database}
\label{T:time-cost}
\begin{center}
\begin{tabular}{c c c c c c}
\hline
 {DynaSLAM} & {SpF} & {SF} & {PF} & {BF} & {Ours} \\
\hline
 1.58 & 0.2- & 16.74 & 0.3- & 24.04(37.89) & 25.84 \\ 
\hline
\end{tabular}
\end{center}
\vspace{-0.5cm }
\end{table}

\subsection{Time Cost Evaluations}
\label{sec:time-cost}
%%%%%%%%%%%%%%%%%%%%%
The online processing \emph{fps} comparison is shown in Tab.~\ref{T:time-cost}. The first four methods use a single GPU and BF uses two GPUs. The \emph{fps} of SpF and PF depend on the situations within the dynamic scenes. For SpF, the frames containing complex non-rigid motions cost more computation time; for PF, the number of human objects and visible body joints result in lower \emph{fps} performance. In the TUM dynamic sequence, SpF is lower than $0.2$ \emph{fps}, and PF is lower than $0.3$ \emph{fps} on average.
DynaSLAM spends more than $500~ms$ on Mask-RCNN 2D human object segmentation, and then applies ORB based VO front end for camera tracking. 
BF performed at $37.89$ \emph{fps} in static scenes, \eg $f1/xyz$, but it drops to $24$ \emph{fps} in a dynamic scene since the back-end loop detector fails when it tries to close the loop in dynamic environments.
%and wastes a lot of computations.
Our method achieved $25.84$ \emph{fps} in the TUM dataset which contains two fast-moving human objects (see Fig.~\ref{fig:airs-trj} (c) and (d)) and $27.79$ \emph{fps} in HRPSlam dataset which contain five moving humans (see Fig.~\ref{fig:hrp-LoopPerformance}).
Our real-time performance in such hard dynamic scenes benefits from speed-up resulting from the usage of four-GPU -- here, we adopted BF's dual-GPU system design as a basic supplemented with a YOLOv4 based dual-BBox human object detector and SMPL model-based human mesh recovery pipeline.   

\section{Conclusions}
In this paper, we present a real-time dense RGB-D SLAM approach for reconstructing accurate static backgrounds together with representing human pose and shape in dynamic humans environment. 
%which simultaneously performs tracking and reconstruction for both rigid backgrounds and the dynamic humans. 
%need to tune this saying:
To the best of our knowledge, this is the first SLAM approach that provides both static backgrounds and dynamic human objects in real-time. 
The whole system runs at 26 \emph{fps} without online GUI rendering (21 \emph{fps} with GUI).
%The dynamic environment reconstructions achieved 21 fps (without human mesh rendering).
%The human pose and shape are recovered as meshes together with their moving trajectories, which are meaningful to HRI applications. 
The represented human mesh is also directly amenable to be deployed for human-robot collaborative loco-manipulation with mobile robots to efficiently reason over target actions in unknown dynamic environments.
Future research directions include involving temporal coherence to refine the estimated 3D human shape and optimizing the loop detector by considering environment semantic information.
%Experimental results on three different datasets indicated that the proposed approach provides not only accurate environment maps but also well-reconstructed moving humans.

\section*{ACKNOWLEDGEMENT}
This work is supported by the Shenzhen Institute of Artificial Intelligence and Robotics for Society (2019-ICP002), The Alan Turing Institute and EU H2020 project Enhancing Healthcare with Assistive Robotic Mobile Manipulation (HARMONY, 9911237).
%%%%%%%%%%%%%%%%%%%%%%%%%%%%%%%%%%%%%%%%%%%%%%%%%%%%%%%%%%%%%%%%%%%%%%%%%%%%%%%%
\bibliographystyle{IEEEtran.bst}
\bibliography{IEEEexample.bib}

\end{document}